\documentclass[]{fairmeta}
\usepackage[numbers]{natbib}
\usepackage{amsthm,amsmath,amssymb}
\usepackage{graphicx}
\usepackage{booktabs}
\usepackage[table]{xcolor}
\usepackage{bm}
\usepackage{xspace}
\usepackage{tabularx}
\usepackage{geometry}
\usepackage{multirow}
\usepackage{enumitem}
\usepackage{placeins}
\usepackage{iftex}
\ifPDFTeX
\else
  \usepackage{fontspec}
\fi
\usepackage{titlesec}
\usepackage{subcaption}
\usepackage[hidelinks,breaklinks,colorlinks=true, linkcolor=blue, citecolor=red, urlcolor=cyan]{hyperref}
\newcolumntype{L}{>{\raggedright\arraybackslash}X}
\newcolumntype{C}{>{\centering\arraybackslash}X}
\newcolumntype{R}{>{\raggedleft\arraybackslash}X}

\ifPDFTeX
  \newcommand{\calmfont}[1]{{#1}}
  \titleformat*{\section}{\Large\bfseries}
  \titleformat*{\subsection}{\large\bfseries}
  \titleformat*{\subsubsection}{\normalsize\bfseries}
\else
  \newfontfamily\calmregular[BoldFont=Outfit-Regular.ttf]{Outfit-Regular.ttf}
  \newcommand{\calmfont}[1]{{{\calmregular #1}}}
  \titleformat*{\section}{\Large\bfseries\calmregular}
  \titleformat*{\subsection}{\large\bfseries\calmregular}
  \titleformat*{\subsubsection}{\normalsize\bfseries\calmregular}
\fi

\newtheorem{theorem}{Theorem}[]

\newtheorem{remark1}[theorem]{Remark}

\newcommand{\modelname}{HoloMotion-1\xspace}

\title{\calmfont{\modelname Technical Report}}

\author{\calmfont{Maiyue Chen}}
\author{\calmfont{Kaihui Wang}}
\author{\calmfont{Bo Zhang}}
\author{\calmfont{Xihan Ma}}
\author{\calmfont{Zhiyuan Yang}}
\author{\calmfont{Yi Ren}}
\author{\calmfont{Qijun Huang}}
\author{\calmfont{Zihao Zhu}}
\author{\calmfont{Yucheng Wang}}
\author{\calmfont{Zhizhong Su}}

\affiliation{\calmfont{Horizon Robotics}}
\code{\url{https://github.com/HorizonRobotics/HoloMotion}}
\webpage{\url{https://horizonrobotics.github.io/robot_lab/holomotion}}
\correspondence{Yucheng Wang at \email{yucheng.wang@horizon.auto}}

\begin{document}

\abstract{
In this report, we present HoloMotion-1, a humanoid motion foundation model for zero-shot whole-body motion tracking. 
A key innovation of HoloMotion-1 is to scale control-policy training with a large-scale hybrid motion corpus, where video-reconstructed motions from in-the-wild videos provide the dominant source of motion diversity, while curated motion-capture and in-house motion data provide higher-fidelity supervision and deployment-oriented coverage. 
This data regime enables HoloMotion-1 to move beyond conventional MoCap-only training and exposes the policy to substantially broader behaviors, capture conditions, and motion styles.

Learning from such heterogeneous data introduces new challenges, including reconstruction noise, source-domain mismatch, uneven motion quality, and the need for temporal modeling under large behavioral variation. 
To address these challenges, HoloMotion-1 integrates large-capacity temporal modeling, a sparsely activated Mixture-of-Experts (MoE) Transformer with KV-cache inference for real-time control, and a sequence-level training strategy that improves learning efficiency on extended motion sequences. 
Extensive experiments on multiple unseen motion benchmarks show that HoloMotion-1 generalizes robustly across diverse motion types and capture conditions, significantly improves tracking accuracy over prior methods, and transfers directly to a real humanoid robot without task-specific fine-tuning.
}

\maketitle


\begin{figure}[b!]
    \centering
    \includegraphics[width=1.0\linewidth]{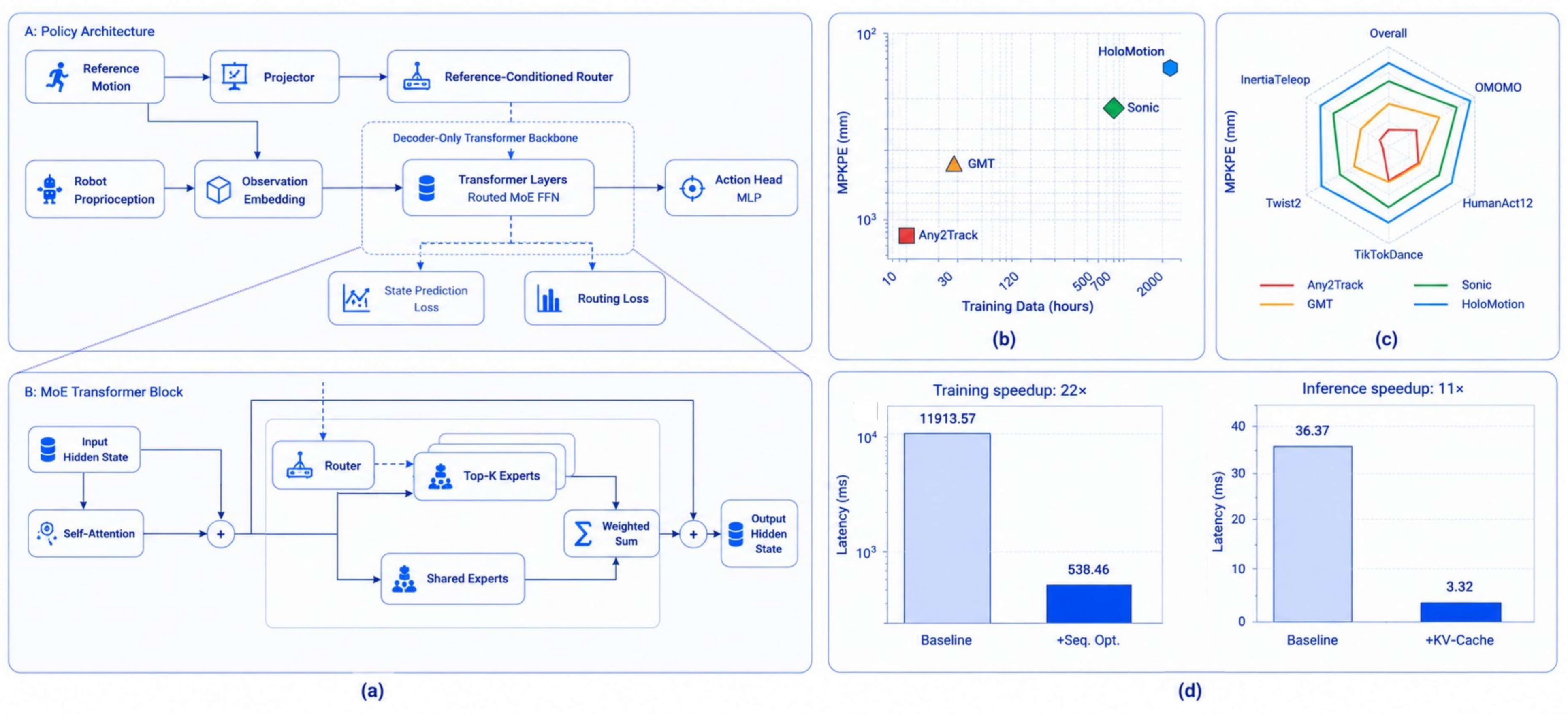}
    \caption{(a) The MoE-Transformer policy network architecture; (b) HoloMotion achieves the lowest overall mean per-keypoint position error; (c) HoloMotion consistently outperforms all compared methods across diverse datasets; (d) the training and inference speedup brought by our sequence-level optimization and KV-cache design.}
    \label{fig:main}
\end{figure}

\section{Introduction}
\label{intro}

Robust humanoid motion tracking is fundamental to reliable whole-body control. By imitating reference motions, tracking policies enable agile and expressive full-body behaviors on real robots. Despite recent progress in imitation-based reinforcement learning~\cite{chen2025gmt, luo2025sonic, zhang2025track, han2025kungfubot2}, most existing systems are trained on relatively limited motion capture datasets using modest-capacity policies, and often struggle when confronted with unseen motion styles or degraded sensing conditions.

In contrast, large-scale pretraining has fundamentally reshaped language modeling, where web-scale data combined with high-capacity models enables generalization far beyond carefully curated datasets~\cite{radford2019language}. This paradigm motivates an analogous question for robotics: \emph{can humanoid motion tracking benefit from scaling both motion data diversity and policy capacity, especially when large-scale video-reconstructed motions are incorporated into the training corpus?} In this report we explore this question using motion reconstructed from large collections of in-the-wild videos~\cite{fan2025go}, which provide orders of magnitude more motion diversity than traditional motion capture datasets.

However, scaling humanoid motion tracking is fundamentally harder than scaling language models.
Zero-shot tracking under unseen behaviors and capture conditions requires broad motion coverage, suggesting a transition from small curated MoCap datasets to large heterogeneous motion corpora. 
In HoloMotion-1, the dominant scaling source comes from motions reconstructed from in-the-wild videos, which provide substantially broader behavioral diversity than conventional motion-capture datasets. 
At the same time, we retain curated MoCap and in-house motion sources to provide cleaner supervision and deployment-relevant motion patterns. 
This hybrid regime is more scalable than MoCap-only training, but also introduces reconstruction noise, source-domain heterogeneity, and uneven motion quality. 
As a result, scaling motion tracking to large-scale heterogeneous motion corpora introduces both a modeling challenge that arises from learning from diverse motion signals with imperfect supervision and a systems challenge that arises from meeting strict latency constraints required for real-time humanoid control.

These challenges motivate a foundation-model approach to humanoid motion learning.
Conventional MLP policies are computationally efficient but lack explicit sequence modeling capability, limiting their ability to represent diverse and long-horizon motion patterns. Transformer architectures~\cite{vaswani2017attention} provide a natural interface for sequence modeling and exhibit favorable scaling behavior in large-data regimes~\cite{radford2019language, brown2020language}, making them a promising candidate for learning from large-scale heterogeneous motion corpora. However, naively scaling Transformers introduces two bottlenecks that are less prominent in typical language modeling settings: training high-capacity sequence models becomes expensive at large data scale, and dense Transformer inference can exceed the latency budget required for real-time closed-loop humanoid control.

In this report we present \textbf{HoloMotion-1}, a humanoid motion foundation model trained on a video-derived hybrid motion corpus for robust zero-shot whole-body motion tracking.
HoloMotion-1 is built to absorb large and heterogeneous motion data while remaining deployable for real-time control on physical humanoid robots. 
HoloMotion-1 brings large-scale video-reconstructed motions into a deployable whole-body tracking system, while retaining curated MoCap and in-house motion sources for fidelity and deployment coverage.
At the model level, the system adopts a Transformer-based motion tracking architecture that explicitly models temporal dependencies in motion sequences. To make large-capacity models deployable in latency-constrained robotic settings, HoloMotion-1 employs a sparsely activated Mixture-of-Experts (MoE) Transformer with KV-cache inference, allowing only a small subset of experts to be activated at each control step while preserving high model capacity. 
At the training level, we introduce a sequence-level PPO optimization paradigm that operates on motion segments rather than individual timesteps, reducing redundant computation on long motion clips and substantially improving training efficiency when learning from large-scale heterogeneous motion corpora.

We evaluate HoloMotion-1 through extensive out-of-domain generalization experiments on multiple unseen motion datasets covering diverse motion types and capture devices, together with comparisons against prior methods, efficiency analysis, and direct transfer experiments on real humanoid hardware.

Our main technical components and findings can be summarized as follows:
\begin{itemize}
    \item We present HoloMotion-1, a large-scale humanoid motion foundation model trained on a video-derived hybrid motion corpus. By integrating large-scale heterogeneous motion data, sparse MoE Transformer modeling, and deployment-oriented training and inference design, HoloMotion-1 achieves approximately 40\% lower global tracking error than the strongest evaluated baseline.
    \item We design a sparsely activated MoE Transformer motion tracker with KV-cache inference that preserves high model capacity while meeting real-time closed-loop control constraints, improving inference efficiency by up to $11\times$.
    \item We introduce a sequence-level PPO training paradigm for autoregressive Transformer policies, reducing redundant computation on long motion clips and improving training efficiency by up to $22\times$.
    \item We establish a comprehensive zero-shot evaluation across five unseen motion datasets spanning diverse motion types and capture devices, demonstrating strong generalization and direct transfer to a real humanoid robot without task-specific fine-tuning.
\end{itemize}

\begin{figure*}[h!]
    \centering
    \includegraphics[width=1.0 \textwidth]{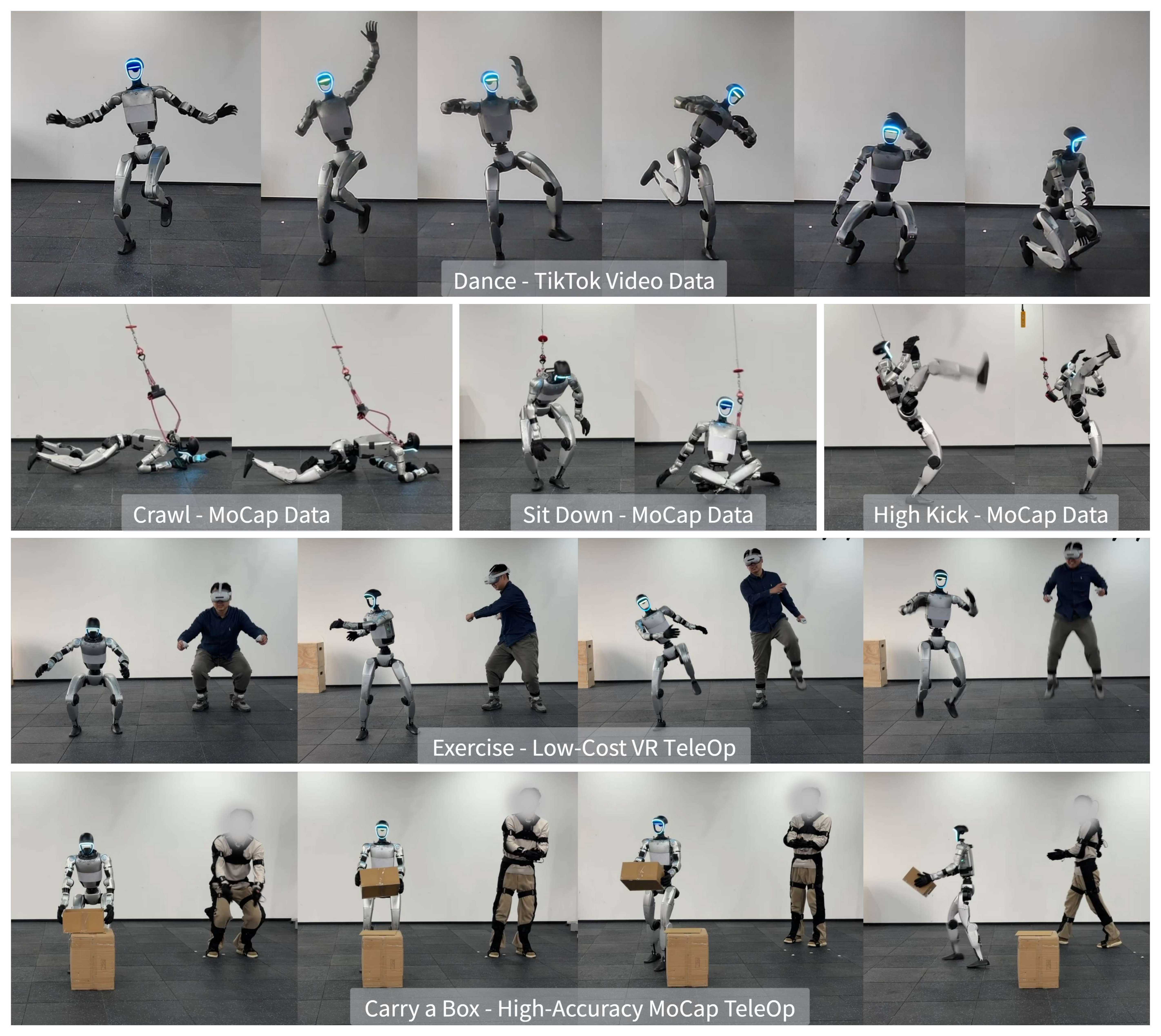}
    \caption{Real-world zero-shot transfer of the HoloMotion policy. In the first row, the robot performs high-dynamic dancing reconstructed from in-the-wild videos. In the second row, the robot performs contact-rich and difficult Kungfu motions. The last two rows showcase real-time teleoperation with robust tracking performance. }
    \label{fig:demo}
\end{figure*}

\section{Background}

\subsection{From Curated Motion Capture to Hybrid Large-Scale Motion Corpora}

Humanoid motion tracking has traditionally relied on curated motion-capture datasets, most notably AMASS~\cite{mahmood2019amass}, as well as laboratory-collected corpora such as LAFAN1~\cite{harvey2020robust}, OMOMO~\cite{li2023object}, and HumanAct12~\cite{guo2020action2motion}. 
These datasets provide clean and well-aligned motion sequences and have enabled stable training of imitation and reinforcement-learning policies. 
However, their scale and behavioral coverage are often limited compared with the diversity required by zero-shot whole-body humanoid tracking.

Recent efforts have started to scale humanoid motion learning with larger motion corpora. 
Some systems scale primarily with high-quality curated or in-house motion data, while another emerging direction reconstructs human motion from large collections of in-the-wild videos~\cite{fan2025go}. 
Video-reconstructed motions provide substantially broader behavioral diversity and are easier to scale, but they also introduce reconstruction artifacts, viewpoint-induced errors, and heterogeneous motion quality. 
HoloMotion-1 follows a hybrid data strategy: it uses video-reconstructed motions as the dominant source of diversity, while complementing them with curated MoCap and in-house motion sources to improve motion fidelity, coverage, and deployment relevance.

\subsection{High-Capacity Policies for Real-Time Humanoid Control}

Humanoid motion tracking policies must balance representational capacity for diverse motions with strict latency requirements for high-frequency closed-loop control. Most existing systems adopt MLP-based policies trained via reinforcement learning~\cite{luo2023perpetual,he2024omnih2o,chen2025gmt,luo2025sonic}. These architectures are computationally efficient and suitable for real-time deployment, but provide limited explicit sequence modeling capability as motion diversity and temporal complexity increase.

To improve policy capacity, some approaches augment MLPs with Mixture-of-Experts (MoE) mechanisms. In many cases, however, MoE layers are deployed with dense activation, which increases both model capacity and per-step computation. Transformer architectures~\cite{vaswani2017attention} provide an alternative sequence modeling paradigm and have demonstrated strong scaling behavior in large-data regimes~\cite{radford2019language,brown2020language}. Transformers have recently been explored in robot manipulation~\cite{zitkovich2023rt,bi2025h} and humanoid control~\cite{fu2024humanplus,tessler2024maskedmimic,radosavovic2024humanoid}. Nevertheless, many existing implementations rely on dense sequence models, where increasing model capacity directly increases inference cost and may exceed the latency budget required for real-time humanoid control.

\subsection{Training High-Capacity Sequence Policies}

Training high-capacity Transformer policies for humanoid control introduces additional challenges, particularly when motion sequences become long. Prior work explores several training paradigms. Some approaches rely on supervised behavior cloning from curated demonstrations~\cite{radosavovic2024humanoid}. Others adopt expert decomposition and distillation strategies to train a generalist Transformer policy~\cite{wang2025experts}. Reinforcement learning approaches such as PPO with history windows are also commonly used to train policies with temporal context~\cite{fu2024humanplus}. 

As model capacity and sequence length increase, however, these training paradigms can become computationally expensive, especially when applied to large-scale motion corpora. Improving training efficiency for sequence policies therefore remains an important challenge when scaling humanoid motion learning to large-scale heterogeneous motion corpora.

\section{HoloMotion-1 Foundation Model}

HoloMotion-1 learns a goal-conditioned control policy for humanoid whole-body motion tracking. 
The control problem is formulated as a goal-conditioned Partially Observable Markov Decision Process (POMDP) defined over a continuous state space $S$, an action space $A$, and an observation space $O$. 
At each control step $t$, the underlying physical state of the robot, including generalized coordinates, velocities, and contact-related variables, is represented by the state vector $\mathbf{s}_t \in S$. 
Because this exact physical state cannot be perfectly measured on real-world humanoid robots, the environment provides an observation vector $\mathbf{o}_t \in O$ generated according to an observation model $p(\mathbf{o}_t \mid \mathbf{s}_t)$.

The tracking objective is specified by a reference kinematic state provided by the environment, denoted as the goal state $\mathbf{s}^{\mathrm{ref}}_t$. 
In practice, this goal specification may include a short-horizon lookahead window $\mathbf{s}^{\mathrm{ref}}_{t:t+H}$, which exposes future reference poses to the policy and allows the controller to anticipate upcoming motion transitions. 
Conditioned on the processed observation or observation history $\tilde{\mathbf{o}}_t$, the policy selects a continuous control action $\mathbf{a}_t \in A$ according to the parameterized policy $\pi_\theta(\cdot \mid \tilde{\mathbf{o}}_t)$. 
After applying this action, the environment evolves to a new state $\mathbf{s}_{t+1}$ according to the transition dynamics distribution $p(\cdot \mid \mathbf{s}_t, \mathbf{a}_t)$.

At every step the agent receives a reward signal measuring tracking performance with respect to the reference motion. 
This reward is defined as $r(\mathbf{s}_t, \mathbf{a}_t; \mathbf{s}^{\mathrm{ref}}_t)$ and captures how closely the robot's motion matches the reference trajectory while maintaining physically stable behavior. 
Given a discount factor $\gamma \in [0,1)$, the learning objective is to optimize the policy parameters $\theta$ such that the expected discounted return

\[
\mathbb{E}\Big[\sum_{t\ge0}\gamma^t r(\mathbf{s}_t,\mathbf{a}_t;\mathbf{s}^{\mathrm{ref}}_t)\Big]
\]

is maximized. 
The complete control formulation can therefore be summarized by the POMDP tuple

\[
P=\langle S,A,p,r,O,\gamma\rangle.
\]

This formulation provides a unified interface for training HoloMotion-1 on large-scale motion corpora while enabling deployment under real-time closed-loop humanoid control.

\subsection{Motion Tracking Formulation}

This section describes the task formulation used to train HoloMotion-1 for humanoid whole-body motion tracking. The formulation defines the observation representation, reward function, termination conditions, domain randomization strategy, and the low-level control interface that together specify the learning problem for the tracking policy.

\subsubsection{Observation}

At each control step $t$, the policy receives an observation vector composed of two primary components: proprioceptive feedback and reference motion features, denoted as $\mathbf{o}^{\mathrm{prop}}_t$ and $\mathbf{o}^{\mathrm{ref}}_t$, respectively. The complete observation vector is defined as

\begin{equation}
\mathbf{o}_t = \big[\mathbf{o}^{\mathrm{prop}}_t,\; \mathbf{o}^{\mathrm{ref}}_t\big].
\end{equation}

The proprioceptive component $\mathbf{o}^{\mathrm{prop}}_t$ captures the currently observable physical state of the robot together with recent control history. This includes the projected gravity vector, root angular velocity, joint positions and velocities, as well as the action applied at the previous control step.

The reference component $\mathbf{o}^{\mathrm{ref}}_t$ encodes the target motion trajectory. It contains the current reference kinematic state $\mathbf{s}^{\mathrm{ref}}_t$ together with a lookahead window spanning $H=10$ future control steps. This lookahead window provides the policy with short-horizon future motion targets, including reference gravity, root velocity, joint targets, and root height, enabling the controller to anticipate upcoming motion transitions.

During training, the actor policy operates on noise-injected observations $\tilde{\mathbf{o}}_t$, which are obtained by adding sampled noise $\boldsymbol{\epsilon}_t$ to the clean observation:

\begin{equation}
\tilde{\mathbf{o}}_t = \mathbf{o}_t + \boldsymbol{\epsilon}_t.
\end{equation}

In contrast, the critic network evaluates the state using the clean observation $\mathbf{o}_t$ augmented with privileged information $\mathbf{o}^{\mathrm{priv}}_t$. This privileged vector includes exact simulation states that are not available on real robots, such as anchor pose differences, heading-aligned reference root states, and precise robot link states. Providing this information during training improves the accuracy of value estimation without affecting deployment. Table~\ref{tab:motrack_obs_terms} summarizes the observation design and the associated noise distributions.

\begin{table}[!htbp]
\centering
\setlength{\tabcolsep}{6pt}
\begin{tabularx}{\linewidth}{L C C}
\toprule
\textbf{Observation} & \textbf{Equation} & \textbf{Actor noise} \\
\midrule
\multicolumn{3}{l}{\textit{Proprioceptive Observation}} \\
\quad Projected gravity & $\mathbf{g}^{\mathrm{proj}}_t$ & $\mathcal{U}(-0.1, 0.1)$ \\
\quad Root angular velocity & $\boldsymbol{\omega}_t$ & $\mathcal{U}(-0.2, 0.2)$ \\
\quad Joint positions (rel.) & $\mathbf{q}_t-\mathbf{q}_0$ & $\mathcal{U}(-0.01, 0.01)$ \\
\quad Joint velocities & $\dot{\mathbf{q}}_t$ & $\mathcal{U}(-0.5, 0.5)$ \\
\quad Previous action & $\mathbf{a}_{t-1}$ & None \\
\midrule
\multicolumn{3}{l}{\textit{Goal Kinematic States}} \\
\quad Ref. projected gravity & $\hat{\mathbf{g}}^{\mathrm{proj}}_{t:t+H}$ & $\mathcal{U}(-0.1, 0.1)$ \\
\quad Ref. base lin. vel & $\hat{\mathbf{v}}^{\mathrm{base}}_{t:t+H}$ & $\mathcal{U}(-0.1, 0.1)$ \\
\quad Ref. base ang. vel & $\hat{\boldsymbol{\omega}}^{\mathrm{base}}_{t:t+H}$ & $\mathcal{U}(-0.1, 0.1)$ \\
\quad Ref. joint targets & $\hat{\mathbf{q}}_{t:t+H}$ & $\mathcal{U}(-0.05, 0.05)$ \\
\quad Ref. root height & $\hat{h}_{t:t+H}$ & $\mathcal{U}(-0.1, 0.1)$ \\
\bottomrule
\end{tabularx}
\caption{Observation design and the noise perturbation added to the actor policy during training.}
\label{tab:motrack_obs_terms}
\end{table}
\FloatBarrier

\subsubsection{Reward}

The reward function is designed to encourage accurate motion tracking while maintaining stable and physically plausible robot behavior. A dense reward is used that combines multiple tracking objectives with regularization terms:

\begin{equation}
r_t = \sum_{i} w_i\, r^{(i)}_t.
\end{equation}

Tracking terms measure how well the robot follows the reference motion through root-relative key-body tracking, root velocity tracking, and a local five-point body-position objective. Absolute root position and orientation rewards are disabled in the configuration used for our main experiments. Additional penalty terms regularize the control policy by discouraging excessive control variation, high joint acceleration, joint limit violations, and unstable contacts. These regularization terms improve training stability and contribute to robust real-world deployment. For ratio rewards, $\rho(\cdot,\cdot)$ denotes the relative velocity error with a small stabilizing constant $\epsilon=0.1$. The local point set $B_5$ contains the torso, both wrists, and both ankles, with a vertical torso offset. Table~\ref{tab:motrack_reward_terms} lists the reward components used in our main experiments.

\begin{table*}[!t]
\centering
\setlength{\tabcolsep}{12pt}
\begin{tabular}{l l r}
\toprule
\textbf{Reward Term} & \textbf{Equation} & \textbf{Weight} \\
\midrule
\multicolumn{3}{l}{\textit{Tracking rewards}} \\
\quad Alive & $r_{\mathrm{alive}}(t)=1$ & 0.1 \\
\quad Key body position & $r^{\mathrm{kb}}_{\mathrm{pos}}(t)=\exp\!\Big(-\frac{1}{|B|}\!\sum_{b\in B}\frac{\|\mathbf{p}^{p,\mathrm{rel}}_{t,b}-\mathbf{p}^{g,\mathrm{rel}}_{t,b}\|_2^2}{0.3^2}\Big)$ & 1.0 \\
\quad Key body rotation & $r^{\mathrm{kb}}_{\mathrm{ori}}(t)=\exp\!\Big(-\frac{1}{|B|}\!\sum_{b\in B}\frac{d_{\mathrm{quat}}(\mathbf{q}^{p,\mathrm{rel}}_{t,b},\mathbf{q}^{g,\mathrm{rel}}_{t,b})^2}{0.4^2}\Big)$ & 1.0 \\
\quad Key body linear velocity & $r^{\mathrm{kb}}_{\mathrm{lin}}(t)=\exp\!\Big(-\frac{1}{|B|}\!\sum_{b\in B}\frac{\|\mathbf{v}^{p}_{t,b}-\mathbf{v}^{g}_{t,b}\|_2^2}{1.0^2}\Big)$ & 1.0 \\
\quad Key body angular velocity & $r^{\mathrm{kb}}_{\mathrm{ang}}(t)=\exp\!\Big(-\frac{1}{|B|}\!\sum_{b\in B}\frac{\|\boldsymbol{\omega}^{p}_{t,b}-\boldsymbol{\omega}^{g}_{t,b}\|_2^2}{3.14^2}\Big)$ & 1.0 \\
\quad Root linear velocity ratio & $r^{\mathrm{root}}_{\mathrm{lin}}(t)=\exp\!\Big(-\frac{\rho(\mathbf{v}^{p}_t,\mathbf{v}^{g}_t)^2}{1.0^2}\Big)$ & 1.0 \\
\quad Root angular velocity ratio & $r^{\mathrm{root}}_{\mathrm{ang}}(t)=\exp\!\Big(-\frac{\rho(\boldsymbol{\omega}^{p}_t,\boldsymbol{\omega}^{g}_t)^2}{1.0^2}\Big)$ & 1.0 \\
\quad Local five-point position & $r^{\mathrm{5pt}}_{\mathrm{pos}}(t)=\exp\!\Big(-\frac{1}{5}\!\sum_{b\in B_5}\frac{\|\mathbf{p}^{p,\mathrm{local}}_{t,b}-\mathbf{p}^{g,\mathrm{local}}_{t,b}\|_2^2}{0.1^2}\Big)$ & 2.0 \\
\midrule
\multicolumn{3}{l}{\textit{Penalty terms}} \\
\quad Action rate & $r_{\mathrm{act}}(t)=\|\mathbf{a}_t-\mathbf{a}_{t-1}\|_2^2$ & -0.2 \\
\quad Joint acceleration & $r_{\mathrm{acc}}(t)=\|\ddot{\mathbf{q}}_t\|_2^2$ & $-10^{-6}$ \\
\quad Joint limit & $r_{\mathrm{jlim}}(t)=\sum_{j} \mathbb{I}\!\big[\mathbf{q}_{t,j} \notin [\mathbf{q}_{t,j}^{\min}, \mathbf{q}_{t,j}^{\max}]\big]$ & -10.0 \\
\quad Undesired contacts & $r_{\mathrm{contact}}(t)=\sum_{c \notin C_{\mathrm{allow}}} \mathbb{I}\!\big[\|\mathbf{F}_{t,c}\|>F_{\mathrm{th}}\big]$ & -0.1 \\
\bottomrule
\end{tabular}
\caption{Reward function formulations and corresponding coefficients used for motion tracking training.}
\label{tab:motrack_reward_terms}
\end{table*}
\FloatBarrier

\subsubsection{Termination}

Episode termination occurs either when the maximum episode length is reached or when the robot deviates significantly from the reference motion. In our main experiments, early termination is defined using projected-gravity, key-body height, and pelvis position thresholds.

First, let $\hat{\mathbf{g}}^{\mathrm{proj}}_t$ denote the reference projected gravity vector. An episode terminates if the Euclidean distance between the robot's projected gravity vector and the reference exceeds $\delta_g=0.8$:

\[
\|\mathbf{g}^{\mathrm{proj}}_t-\hat{\mathbf{g}}^{\mathrm{proj}}_t\|_2 > 0.8 .
\]

Second, let $B_z$ denote critical body links consisting of the pelvis, both ankles, and both wrists. The episode terminates if the maximum vertical tracking error across these links exceeds $\delta_z=0.25$:

\[
\max_{b\in B_z}
|p^{p,z}_{t,b}-p^{g,z}_{t,b}| > 0.25 .
\]

Third, pelvis position drift is constrained directly:

\[
\|\mathbf{p}^{p}_{t,\mathrm{pelvis}}-\mathbf{p}^{g}_{t,\mathrm{pelvis}}\|_2 > 0.25 .
\]

These termination conditions prevent unstable rollouts and encourage policies that maintain consistent alignment with the reference motion.

\subsubsection{Domain Randomization}

To improve robustness and sim-to-real transfer, domain randomization is applied during training. The configuration used for our main experiments randomizes observation noise, action delay, initial motion state, contact material properties, mass, center-of-mass offsets, and actuator gains. In addition, periodic external perturbations in the form of random pushes are applied during rollouts, and training is performed on a rough height-field terrain. These perturbations expose the policy to disturbances and encourage recovery behaviors. Table~\ref{tab:domain_rand} summarizes the domain randomization configuration.

\begin{table}[!htbp]
\centering
\setlength{\tabcolsep}{10pt}
\begin{tabular}{l r}
\toprule
\textbf{Parameter} & \textbf{Sampling Distribution} \\
\midrule
\multicolumn{2}{l}{\textit{Control and terrain}} \\
\quad Action delay & $d_a \sim \mathrm{Uniform}\{0,1,2\}$ steps \\
\quad Terrain height noise & $h_{\mathrm{terrain}}\sim U[0.0,0.04]\ \mathrm{m}$ \\
\midrule
\multicolumn{2}{l}{\textit{Physical parameters}} \\
\quad Static friction coefficient & $\mu_s \sim U[0.3, 1.6]$ \\
\quad Dynamic friction coefficient & $\mu_d \sim U[0.3, 1.2]$ \\
\quad Restitution coefficient & $e \sim U[0, 0.5]$ \\
\quad Default joint positions & $q_0 \leftarrow q_0 + U[-0.01, 0.01]$ \\
\quad Pelvis/torso mass offset & $\Delta m \sim U[-1.0, 2.0]\ \mathrm{kg}$ \\
\multirow{3}{*}{\quad Torso COM offset} & $\Delta x \sim U[-0.075, 0.075]$ \\
& $\Delta y \sim U[-0.1, 0.1]$ \\
& $\Delta z \sim U[-0.1, 0.1]$ \\
\quad PD gain scaling & $k_p, k_d \leftarrow k_p, k_d \cdot U[0.9, 1.1]$ \\
\midrule
\multicolumn{2}{l}{\textit{Initial-state perturbations}} \\
\quad Root position & $\Delta x,\Delta y\sim U[-0.05,0.05],\ \Delta z\sim U[-0.01,0.01]$ \\
\quad Root orientation & $\Delta r,\Delta p\sim U[-0.1,0.1],\ \Delta \psi\sim U[-0.2,0.2]$ \\
\quad Root velocity & $\Delta v_{x,y}\sim U[-0.5,0.5],\ \Delta v_z\sim U[-0.2,0.2]$ \\
\quad Joint positions & $\Delta q \sim U[-0.1,0.1]$ \\
\midrule
\multicolumn{2}{l}{\textit{Root velocity perturbations}} \\
\quad Push interval & $\Delta t \sim U[1, 3]$ \\
\multirow{3}{*}{\quad Root linear velocity} & $v_x \sim U[-0.5, 0.5]$ \\
& $v_y \sim U[-0.5, 0.5]$ \\
& $v_z \sim U[-0.2, 0.2]$ \\
\multirow{3}{*}{\quad Root angular velocity} & $\omega_{\mathrm{roll}} \sim U[-0.52, 0.52]$ \\
& $\omega_{\mathrm{pitch}} \sim U[-0.52, 0.52]$ \\
& $\omega_{\mathrm{yaw}} \sim U[-0.78, 0.78]$ \\
\bottomrule
\end{tabular}
\caption{Domain randomization parameters and their corresponding sampling distributions applied during training.}
\label{tab:domain_rand}
\end{table}
\FloatBarrier

\subsubsection{Action and Low-Level Control}

The policy outputs normalized action commands that parameterize joint position targets around default joint configurations:

\begin{equation}
\mathbf{q}^{\mathrm{tar}}_t = \mathbf{q}_0 + \mathbf{s}\odot \mathbf{a}_t,
\end{equation}

where $\mathbf{s}$ denotes a per-joint scaling vector.

These target positions are tracked by per-joint proportional-derivative (PD) controllers that convert position errors into torque commands:

\begin{equation}
\boldsymbol{\tau}_t
=
\mathbf{k}_p \odot (\mathbf{q}^{\mathrm{tar}}_t-\mathbf{q}_t)
-
\mathbf{k}_d \odot \dot{\mathbf{q}}_t.
\end{equation}

This control interface provides a stable low-level actuation layer while allowing the learned policy to operate in joint position space.

\subsection{Sparse MoE Transformer Policy}

HoloMotion-1 adopts a causal decoder-only Transformer architecture with sparsely activated Mixture-of-Experts (MoE) layers to map goal-conditioned observation streams to continuous joint-space control actions. 
The control policy is implemented as a Transformer sequence model that processes temporal observation tokens and outputs joint-level control commands. 
During training, a separate value network is used to estimate the expected return of the current state using noise-free observations and privileged information. 
Figure~\ref{fig:main}-(a) illustrates the overall architecture of the HoloMotion-1 policy network.

\subsubsection{Tokenization}

At each control step $t$, the observation is converted into a token representation $\mathbf{x}_t \in \mathbb{R}^{d_{\mathrm{in}}}$ by concatenating proprioceptive signals and reference motion features. 
The reference component includes a fixed look-ahead window of $H$ future reference frames (see Table~\ref{tab:motrack_obs_terms}), allowing the model to anticipate upcoming motion transitions.

The resulting observation vector is normalized using a running empirical mean and variance updated via an exponential moving average (EMA) during training. 
The normalized vector is then projected into the hidden model dimension $d$ through a lightweight MLP projection layer to form the Transformer input token.

\subsubsection{Transformer Backbone}

The core control model is a decoder-only Transformer~\cite{vaswani2017attention} that operates autoregressively over a rolling context window of the most recent $C$ tokens. 
The architecture follows a pre-norm residual design with RMSNorm~\cite{zhang2019root} and rotary positional embeddings (RoPE)~\cite{su2024roformer}.

To improve computational efficiency, the causal self-attention mechanism uses grouped-query attention (GQA)~\cite{ainslie2023gqa}, configured with $n_h$ query heads and $n_{\mathrm{kv}}$ key-value heads. 
Training stability is further enhanced using query-key normalization (QK-Norm)~\cite{henry2020query} and a lightweight sigmoid gating mechanism applied to attention outputs (Gated-Attention)~\cite{qiu2025gated}.

This Transformer backbone enables explicit sequence modeling of motion trajectories while maintaining efficient inference for real-time control.

\subsubsection{Mixture-of-Experts}

To increase model capacity while maintaining bounded inference cost, HoloMotion-1 employs a sparse Mixture-of-Experts (MoE) architecture. 
The policy uses a reference-routed grouped MoE Transformer block without a leading dense feed-forward layer. This design concentrates capacity in a large expert pool while keeping the number of activated experts small at each control step.

For an input token representation $\mathbf{u}$, the MoE layer combines a shared dense expert $f_{\mathrm{sh}}$ with a set of $E$ specialized experts $\{f_i\}_{i=1}^{E}$. 
A learned routing network selects the top-$k$ experts for each token, forming the index set $T_k(\mathbf{u})$. 
Each selected expert contributes to the output using normalized mixture weights $\alpha_i(\mathbf{u})$. 
The resulting computation is

\begin{equation}
\mathrm{MoE}(\mathbf{u}) = f_{\mathrm{sh}}(\mathbf{u}) + \sum_{i\in T_k(\mathbf{u})} \alpha_i(\mathbf{u})\, f_i(\mathbf{u}).
\end{equation}

This sparse expert routing significantly increases the model's representational capacity while limiting the per-token expert computation to $k$, making the architecture suitable for real-time humanoid control.

\subsubsection{Router}
Routing quality is a key factor in the effectiveness of MoE policies. We find that naively combining proprioceptive observations with reference motion inputs for routing leads to severe sim-to-real degradation. This occurs because routing decisions significantly influence policy behavior; when the router becomes sensitive to low-level dynamic variations, unavoidable sim-to-real discrepancies can amplify routing oscillations and ultimately destabilize real-world execution. Therefore, we deliberately separate the router input pathway and condition the router exclusively on the reference motion input. This design reduces the sensitivity of expert assignment to robot-state sim-to-real discrepancies, while keeping routing decisions consistent for the same reference motion across simulation and deployment.

\subsubsection{Action Distribution}

The final Transformer hidden state is mapped to a Gaussian action distribution using a lightweight MLP prediction head. 
The head outputs the mean vector $\boldsymbol{\mu}_t$ of the policy distribution, while the policy uncertainty is parameterized by a learnable state-independent log-standard-deviation vector $\log\boldsymbol{\sigma}$, forming a diagonal covariance matrix $\mathrm{diag}(\boldsymbol{\sigma}^2)$. 
The continuous action $\mathbf{a}_t$ is sampled from this Gaussian distribution.

\subsubsection{Auxiliary Losses}
To improve optimization efficiency and encourage the Transformer to encode motion-relevant structure before sparse expert routing, we attach auxiliary prediction heads to the pre-MoE hidden representation. These heads are used only during training and provide direct supervision on low-dimensional kinematic signals that are strongly correlated with successful motion tracking. In the configuration used for our experiments, the active auxiliary objectives supervise base linear velocity, key-body contact states, reference key-body positions relative to the root, and robot key-body positions relative to the root. We additionally employ a router-side dead-expert margin regularizer to mitigate expert under-utilization.

Let $m_{bt} \in \{0,1\}$ denote the valid-token mask for batch index $b$ and time index $t$, and let $N = \sum_{b,t} m_{bt}$. The auxiliary objective added to the actor loss is
$$
\mathcal{L}_{\mathrm{aux}}
=
\lambda_{\mathrm{vel}} \mathcal{L}_{\mathrm{vel}}
+
\lambda_{\mathrm{contact}} \mathcal{L}_{\mathrm{contact}}
+
\lambda_{\mathrm{ref}} \mathcal{L}_{\mathrm{ref}}
+
\lambda_{\mathrm{robot}} \mathcal{L}_{\mathrm{robot}}
+
\lambda_{\mathrm{dead}} \mathcal{L}_{\mathrm{dead}},
$$
where the released HoloMotion-1 configuration uses $\lambda_{\mathrm{vel}}=\lambda_{\mathrm{contact}}=10^{-2}$, $\lambda_{\mathrm{ref}}=\lambda_{\mathrm{robot}}=10^{-1}$, and $\lambda_{\mathrm{dead}}=10^{-1}$.

For base-velocity prediction, the auxiliary head outputs a diagonal Gaussian with mean $\hat{\boldsymbol{\mu}}^{v}_{bt} \in \mathbb{R}^{3}$ and standard deviation $\hat{\boldsymbol{\sigma}}^{v}_{bt} \in \mathbb{R}^{3}$. The corresponding negative log-likelihood is
$$
\mathcal{L}_{\mathrm{vel}}
=
\frac{1}{N}
\sum_{b,t} m_{bt}
\cdot
\frac{1}{2}
\sum_{j=1}^{3}
\left[
\left(
\frac{v_{btj} - \hat{\mu}^{v}_{btj}}
{\hat{\sigma}^{v}_{btj}}
\right)^{2}
+
2 \log \hat{\sigma}^{v}_{btj}
\right],
$$
where $\mathbf{v}_{bt}$ is the ground-truth base linear velocity and the predicted standard deviation is clamped to a fixed interval for numerical stability. This term encourages the pre-MoE representation to preserve compact global dynamics information.

Let $K_c$ denote the number of supervised key bodies for contact prediction, let $c_{btk} \in \{0,1\}$ be the ground-truth contact label, and let $\hat{\ell}_{btk}$ be the predicted contact logit. The contact loss is
$$
\mathcal{L}_{\mathrm{contact}}
=
\frac{1}{N K_c}
\sum_{b,t} m_{bt}
\sum_{k=1}^{K_c}
\mathrm{BCE}\bigl(\hat{\ell}_{btk}, c_{btk}\bigr),
$$
where $\mathrm{BCE}(\cdot,\cdot)$ denotes the binary cross-entropy with logits. This objective promotes sensitivity to gait phase and support transitions.

Let $K_p$ denote the number of supervised key bodies for position prediction. The reference-position head predicts root-relative reference key-body positions $\hat{\mathbf{p}}^{\mathrm{ref}}_{btk} \in \mathbb{R}^{3}$ and is supervised by $\mathbf{p}^{\mathrm{ref}}_{btk}$ from the current reference motion. The corresponding masked mean-squared error is
$$
\mathcal{L}_{\mathrm{ref}}
=
\frac{1}{3 N K_p}
\sum_{b,t} m_{bt}
\sum_{k=1}^{K_p}
\left\|
\hat{\mathbf{p}}^{\mathrm{ref}}_{btk}
-
\mathbf{p}^{\mathrm{ref}}_{btk}
\right\|_2^2.
$$

The robot-position head uses the same supervised body set, but its targets are the robot key-body positions transformed into the root-relative frame. If $\hat{\mathbf{p}}^{\mathrm{robot}}_{btk}, \mathbf{p}^{\mathrm{robot}}_{btk} \in \mathbb{R}^{3}$ are the predicted and target robot key-body positions, the loss is

$$
\mathcal{L}_{\mathrm{robot}}
=
\frac{1}{3 N K_p}
\sum_{b,t} m_{bt}
\sum_{k=1}^{K_p}
\left\|
\hat{\mathbf{p}}^{\mathrm{robot}}_{btk}
-
\mathbf{p}^{\mathrm{robot}}_{btk}
\right\|_2^2.
$$
Together, these two position losses encourage the pre-MoE representation to encode both the reference motion geometry and the robot-specific kinematic realization.

Finally, to reduce expert collapse in sparse routing, we introduce a dead-expert margin loss. For each MoE layer $\ell$, let $E$ denote the number of fine experts, let $s^{\ell}_{bte}$ be the router score for expert $e$ at token $(b,t)$, and let $\tau^{\ell}_{bt}$ be the score of the $k$-th selected expert, i.e., the routing threshold induced by top-$k$ selection. Let $D_\ell$ be the set of experts that receive no tokens in the current batch. The per-layer dead-expert margin loss is
$$
\mathcal{L}^{\ell}_{\mathrm{dead}}
=
\frac{1}{BT\max\left(1, |D_\ell|\right)}
\sum_{b,t}\sum_{e\in D_\ell}
\left[
\tau^{\ell}_{bt} - s^{\ell}_{bte}
\right]_{+},
$$
where $[x]_{+} = \max(x, 0)$ and $BT$ is the number of tokens in the update batch. The final $\mathcal{L}_{\mathrm{dead}}$ is averaged over MoE layers. This regularizer drives inactive experts toward the current routing frontier, improving expert utilization while preserving the sparsity structure of top-$k$ routing.

\subsubsection{Value Network}

During training, a value network is used to estimate the expected return of the current state. 
The value function is parameterized as an MLP that predicts a scalar value $V_\psi(\mathbf{o}_t, \mathbf{o}^{\mathrm{priv}}_t)$ from the observation and privileged state information.

To improve value estimation accuracy, the value network receives noise-free observations together with privileged simulation states, including anchor pose differences and exact link states. 
These privileged signals are used only during training and are not required during deployment.

In the default HoloMotion-1 configuration, the hidden model dimension is set to $d=512$, and the Transformer contains a single sparse MoE layer without a leading dense feed-forward layer. 
The attention module uses $n_h=8$ query heads and $n_{\mathrm{kv}}=4$ key-value heads with a context window of $C=32$ tokens. 
The MoE routing layer contains $E=1024$ fine experts, includes one shared expert, and activates the top $k=2$ fine experts per token.

\subsection{Sequence-Level Policy Optimization}

Optimizing high-capacity sequence models for humanoid control presents a significant efficiency challenge. 
Standard Proximal Policy Optimization (PPO)~\cite{schulman2017proximal} typically flattens rollout trajectories and updates the policy using randomly shuffled individual time steps. 
This design is well suited for policies that operate on static per-step feature vectors. 
However, HoloMotion-1 employs a causal Transformer policy whose decision at step $t$ depends on an ordered sequence of historical observations up to the context length $C$.

Let the observation prefix be denoted as $\tilde{\mathbf{o}}_{t-C+1:t}$, and the policy as $\pi_\theta(\mathbf{a}_t \mid \tilde{\mathbf{o}}_{t-C+1:t})$. 
Applying standard step-level PPO in this setting requires storing and recomputing the entire observation prefix for every sampled step during optimization. 
Because these prefixes overlap heavily within a mini-batch, this results in redundant $O(C)$ computation per step and becomes a major memory and compute bottleneck when scaling to large motion corpora.

To address this limitation, HoloMotion-1 adopts a \emph{sequence-level policy optimization} strategy. 
Instead of shuffling individual time steps, training preserves contiguous rollout segments of length $T$ collected from a batch of $B$ parallel environments. 
This design leverages the autoregressive factorization of the trajectory distribution:

\begin{equation}
\pi_\theta(\mathbf{a}_{1:T} \mid \tilde{\mathbf{o}}_{1:T})
=
\prod_{t=1}^T
\pi_\theta(\mathbf{a}_t \mid \tilde{\mathbf{o}}_{1:t}).
\end{equation}

With this formulation, a single batched forward pass can compute the action log-probabilities for all $BT$ time steps simultaneously. 
This eliminates the repeated evaluation of highly overlapping observation prefixes and significantly improves training efficiency for long motion sequences.

The resulting optimization objective follows the PPO clipped surrogate loss. 
Let $\theta$ denote the policy parameters, $\hat{A}_t$ the estimated advantage at time step $t$, and $\epsilon$ the PPO clipping parameter. 
The importance ratio $r_t(\theta)$ is defined as the ratio between the action probability under the current policy and the behavior policy used during rollout collection. 
The policy objective is given by

\begin{equation}
L_{RL}(\theta)
=
\hat{\mathbb{E}}_t
\Big[
\min
\big(
r_t(\theta)\hat{A}_t,
\mathrm{clip}(r_t(\theta),1-\epsilon,1+\epsilon)\hat{A}_t
\big)
\Big].
\end{equation}

To support efficient autoregressive inference during rollout collection, HoloMotion-1 maintains a per-environment key-value (KV) cache for the Transformer attention layers. 
The cache is implemented as a ring buffer of length $C$ storing previously computed key-value states. 
At each control step, the model processes only the newly observed token while attending to the cached context, reducing the attention computation complexity from $O(C^2)$ to $O(C)$.

When an episode terminates, the corresponding environment cache is cleared so that the next episode begins with an empty context. 
The same KV-caching mechanism is also used during real-world deployment, enabling the Transformer policy to operate at high control frequencies.

\section{HoloMotion-1 System Pipeline}
\label{sec:pipeline}

\begin{figure*}[t]
    \centering
    \includegraphics[width=\textwidth]{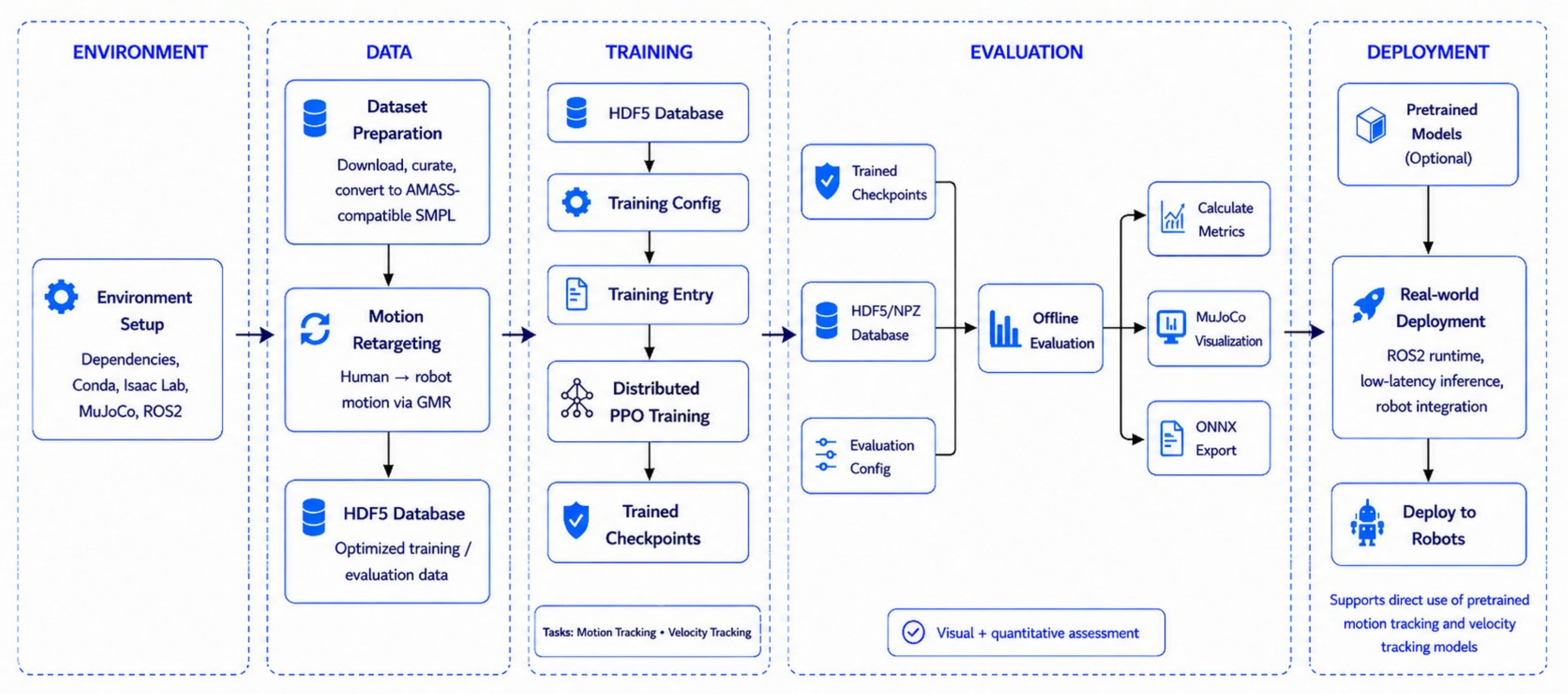}
    \caption{
    The HoloMotion system pipeline. 
    The framework provides an end-to-end workflow covering environment setup, dataset preparation, motion retargeting, distributed PPO training, offline evaluation, policy export, and real-world deployment. 
    This pipeline supports both training from customized motion datasets and direct deployment using pretrained motion tracking and velocity tracking models.
    }
    \label{fig:pipeline}
\end{figure*}

Beyond the model architecture and training algorithm, HoloMotion-1 is designed as an end-to-end system for scalable humanoid motion learning, evaluation, and deployment. 
As shown in Figure~\ref{fig:pipeline}, the system pipeline covers the complete workflow from environment setup and motion data preparation to distributed policy training, offline evaluation, model export, and real-world deployment. 
This design makes HoloMotion not only a motion tracking policy, but also a reproducible framework for developing, evaluating, and deploying whole-body humanoid control models. 
The pipeline is organized around explicit intermediate artifacts, including AMASS-style motion files, retargeted robot trajectories, packed HDF5 databases, trained checkpoints, exported deployment policies, and deployment-ready motion clips.

\textbf{Environment and data preparation.}
The pipeline begins with documented environment setup for simulation, motion processing, training, evaluation, and deployment. 
Motion data from different sources are first converted into AMASS-style SMPL or SMPL-X motion files, and then retargeted from human motion to the target humanoid robot through GMR-based motion retargeting. 
The retargeted trajectories are converted into HoloMotion-compatible NPZ files, which can be visualized in MuJoCo for quality inspection before training. 
For large-scale training and IsaacLab evaluation, these trajectories are further packed into HDF5 databases, providing a compact and efficient format for batched motion loading.

\textbf{Distributed policy training.}
Given the prepared motion database, HoloMotion supports distributed PPO training for motion tracking and velocity tracking tasks. 
Training is configured through modular training files, allowing datasets, robot-specific parameters, observation schemas, reward settings, domain randomization settings, terrain settings, and policy architectures to be reused within the same framework. 
This configuration structure keeps the learning algorithm, robot description, environment definition, and network architecture decoupled, which makes it easier to reproduce released models or adapt the pipeline to new motion datasets. 
The output of this stage is a set of trained checkpoints together with the resolved training configuration required for evaluation and export.

\textbf{Offline evaluation and model export.}
To ensure reproducible benchmarking, HoloMotion provides two complementary offline evaluation paths. 
The IsaacLab evaluation path runs trained checkpoints on HDF5 motion datasets and can dump rollout results as NPZ files. 
The MuJoCo sim-to-sim evaluation path consumes exported policies and NPZ motion datasets, supporting batch evaluation, video rendering, per-clip metrics, and failure-case inspection. 
This separation mirrors the deployment flow: PyTorch checkpoints are used during training and IsaacLab evaluation, while exported policies are used for MuJoCo validation and robot-side runtime deployment.

\textbf{Real-world deployment.}
Finally, HoloMotion includes a deployment workflow for running pretrained or newly trained models on physical humanoid robots. 
The deployment stack includes ROS2 runtime integration, robot-side observation construction, policy inference, and low-level control interfaces. 
The robot runtime supports dual-policy execution: a velocity tracking policy provides locomotion and safe mode transitions, while a motion tracking policy executes either offline NPZ motion clips or live teleoperation references. 
This enables users to either train customized policies from their own motion datasets or directly use released pretrained models for real-world motion tracking and velocity tracking.

Overall, this system pipeline complements the HoloMotion-1 model by providing the infrastructure required to scale humanoid motion learning in practice. 
It standardizes the full lifecycle from motion data construction to real-robot execution, making HoloMotion a practical foundation for future extensions in command following, terrain-aware control, and cross-embodiment deployment.

\section{Experiments}
\subsection{Experimental Setup}

All experiments are conducted on the Unitree G1 humanoid robot platform with 29 degrees of freedom (DoF). 
Policy training is performed in IsaacLab, while evaluation rollouts are executed in the MuJoCo simulator to ensure consistent physics benchmarking across methods.

\subsubsection{Training Corpus}

HoloMotion-1 is trained on a large-scale video-derived hybrid motion corpus that combines reconstructed motions from in-the-wild videos with curated MoCap datasets and in-house motion sources. 
For the video-derived portion, we primarily use MotionMillion~\cite{fan2025go}, an external source of motions reconstructed from in-the-wild monocular videos using SMPL-based motion estimation, to provide scale and behavioral diversity. 
Curated MoCap data such as AMASS and LAFAN1 provides cleaner and more physically consistent motion supervision, while in-house motion data improves coverage of deployment-relevant behaviors such as teleoperation and high-dynamic demonstrations.
The in-house subset is collected with two complementary capture setups: a PICO 4 Ultra VR/MR headset for VR-style teleoperation motions and a Noitom PN Link system for anti-magnetic inertial MoCap and continuous full-body motion capture.

Overall, the full training corpus contains over 2,000 hours of motion data. 
This hybrid data design differs from conventional MoCap-only training: it preserves the scalability and diversity of video-reconstructed motions while mitigating their reconstruction noise and domain heterogeneity through higher-fidelity and deployment-oriented motion sources.
Table~\ref{tab:training_corpus} summarizes the main components of the training corpus.

\begin{table}[!htbp]
\begin{tabularx}{\linewidth}{l X X r}
\toprule
Component & Source / Device & Role & Duration \\ \midrule
Video-derived & MotionMillion~\cite{fan2025go}  & Diversity at scale & 2,000+ hours \\
Curated MoCap & AMASS~\cite{mahmood2019amass} & Clean MoCap & 40+ hours \\
Curated MoCap & LAFAN1~\cite{harvey2020robust} & Studio MoCap & 4.6 hours \\
In-house & PICO 4 Ultra VR/MR headset & Teleoperation & 10+ hours \\
In-house & Noitom PN Link system & Inertial MoCap & 10+ hours \\
\bottomrule
\end{tabularx}
\caption{Training corpus components used by HoloMotion-1. The video-derived portion provides large-scale behavioral diversity, while curated MoCap and in-house sources improve fidelity and deployment relevance.}
\label{tab:training_corpus}
\end{table}
\FloatBarrier

\subsubsection{Training Configuration}

All HoloMotion-1 models are trained on a cluster of 64 NVIDIA RTX 5090 GPUs for approximately six days, corresponding to roughly 9,200 GPU hours in total. 
Training is conducted using large-scale parallel simulation environments in IsaacLab, where 8,192 environments run simultaneously to collect on-policy rollouts for reinforcement learning.

The Transformer-based policy and value networks are optimized using the proposed sequence-level PPO algorithm. 
During training, policy updates are performed on batched trajectory segments collected from multiple parallel environments, enabling efficient learning from long motion sequences.

\subsubsection{Evaluation Datasets}

To comprehensively evaluate the generalization capability of HoloMotion-1, we construct an out-of-domain evaluation suite designed to test motion tracking under diverse motion styles and noise conditions.

To evaluate tracking performance on high-quality reference motions, we use the OMOMO~\cite{li2023object} dataset, which provides precise kinematic trajectories captured using optical motion capture systems.
To assess performance on motions reconstructed from commodity sensors, we use HumanAct12~\cite{guo2020action2motion}, which contains motions extracted from RGB-D recordings.
To evaluate teleoperation-style motion inputs, we include Twist2~\cite{ze2025twist2}, an open dataset designed for VR-based teleoperation scenarios.

In addition, we construct two in-house datasets to further stress-test the model under challenging motion conditions. 
\textbf{TikTokDance} is curated from in-the-wild dance videos and contains highly dynamic and stylistically diverse motion sequences reconstructed from in-the-wild videos. 
\textbf{InertialTeleop} is collected using an inertial motion capture suit and provides real-time teleoperation motion inputs with natural human variability.

Together, these datasets form a heterogeneous evaluation suite covering multiple motion sources and sensing modalities, enabling a comprehensive assessment of zero-shot motion tracking performance.
Table~\ref{tab:eval_sets} summarizes the details of each dataset.

\begin{table}[!htbp]
\begin{tabularx}{\linewidth}{l X X r}
\toprule
Sub-dataset & Device & Activities & Duration \\ \midrule
OMOMO           & Optical MoCap                 & Object interactions   & 9 hours   \\
HumanAct12      & RGB-D cameras           & Daily actions                 & 1.6 hours \\
Twist2          & VR headset          & Teleoperation     & 0.28 hours\\
TikTokDance     & RGB camera        & Popular dance                 & 1 hour   \\
InertialTeleop & Inertial MoCap        &  Teleoperation      & 0.6 hours \\
\bottomrule
\end{tabularx}
\caption{Out-of-domain evaluation sub-datasets spanning diverse activities and capture devices. }
\label{tab:eval_sets}
\end{table}
\FloatBarrier

\subsubsection{Evaluation Metrics}

To mitigate the impact of sample size imbalance across the evaluation datasets, all reported metrics are macro-averaged across the sub-evaluation sets.

Tracking performance is evaluated using four metrics:

\begin{itemize}
    \item \textbf{Mean per-keypoint position error ($E_{\mathrm{mpkpe}}$).} 
    Measured in millimeters (mm), this metric quantifies the global positional deviation of the robot body links relative to the reference motion.

    \item \textbf{Mean per-joint position error ($E_{\mathrm{mpjpe}}$).} 
    Measured in radians (rad), this metric evaluates the tracking error in joint configuration space.

    \item \textbf{Root velocity error ($E_{\mathrm{vel}}$).} 
    Measured in millimeters per frame (mm/frame), this metric assesses the alignment between predicted and reference root velocities.

    \item \textbf{Success rate (SR).} 
    Defined as the percentage of evaluation clips in which the robot root height remains within $0.25\,\mathrm{m}$ of the reference trajectory throughout the entire rollout.
\end{itemize}

Among these metrics, $E_{\mathrm{mpkpe}}$ serves as the primary evaluation metric since it simultaneously reflects both global root motion accuracy and local articulation fidelity.

\subsection{Evaluation Protocol}

The experimental evaluation is designed to assess HoloMotion-1 as a complete system along three key dimensions:

\begin{itemize}
    \item \textbf{Comparison with prior methods:} 
    Performance is compared against existing humanoid motion tracking systems on the same out-of-domain evaluation suite.

    \item \textbf{Training and inference efficiency:} 
    The efficiency improvements introduced by sequence-level optimization and sparse MoE inference are measured.

    \item \textbf{Real-world deployment:} 
    The ability of the learned policy to transfer to real humanoid hardware without task-specific fine-tuning is evaluated.
\end{itemize}

The evaluation focuses on the released HoloMotion-1 system and compares it with representative open-source humanoid motion tracking baselines under a consistent MuJoCo evaluation suite. Whenever available, baseline checkpoints are taken from the authors' official repositories to reduce implementation-specific discrepancies. This protocol highlights the end-to-end capability of HoloMotion-1 as a practical foundation model for zero-shot whole-body humanoid motion tracking.

\subsection{Comparison with Prior Methods}

The performance of HoloMotion-1 is compared with several recent humanoid motion tracking systems, including GMT~\cite{chen2025gmt}, Any2Track~\cite{zhang2025track}, and Sonic~\cite{luo2025sonic}. 
These methods represent recent approaches to large-scale humanoid motion tracking using reinforcement learning and imitation-based control policies.

All methods are evaluated under the same simulation conditions and across the same out-of-domain evaluation datasets. 
For each method, policies are executed in MuJoCo using identical robot models and a consistent evaluation protocol to improve comparability.
Table~\ref{tab:full_results} summarizes the quantitative comparison, and Figure~\ref{fig:main}-(b) and (c) visualize the overall and per-dataset results.

\begin{table*}[!t]
  \centering
  \scriptsize
  \setlength{\tabcolsep}{3.5pt}
  \renewcommand{\arraystretch}{1.1}
  \resizebox{\textwidth}{!}{%
  \begin{tabular}{l | c c | c c c | c c c c}
    \toprule
    \textbf{Method} & \textbf{Training Corpus} & \textbf{Dur.} & \textbf{Architecture} & \textbf{Activated} & \textbf{Total} & $\mathbf{E_{\mathrm{mpkpe}}}$ & $\mathbf{E_{\mathrm{mpjpe}}}$ & $\mathbf{E_{\mathrm{vel}}}$ & \textbf{SR} \\
    \midrule
    GMT       & LAFAN1 \& AMASS & 30 hours   & MLP & 2M  & 2M  & 484.05  & 0.1097 & 8.0  & 92.54\% \\
    Any2Track & LAFAN1          & 10 hours   & MLP & 33M & 33M & 1200.96 & 0.1342 & 13.8 & 25.10\% \\
    Sonic     & In-house        & 700 hours  & MLP & 42M & 42M & 227.95  & 0.1069 & 5.4  & 95.58\% \\
       \midrule
    \multirow{3}{*}{HoloMotion}
      & Video-derived motions
      & \multirow{3}{*}{2000+ hours}
      & \multirow{3}{*}{MoE Transformer}
      & \multirow{3}{*}{7M}
      & \multirow{3}{*}{400M}
      & \multirow{3}{*}{\textbf{124.57}}
      & \multirow{3}{*}{\textbf{0.0979}}
      & \multirow{3}{*}{\textbf{3.9}}
      & \multirow{3}{*}{\textbf{97.55\%}} \\
      & + AMASS/LAFAN1
      &
      &
      &
      &
      &
      &
      &
      &
      \\
      & + In-house motions
      &
      &
      &
      &
      &
      &
      &
      &
      \\
    \bottomrule
  \end{tabular}%
  }
  \caption{Evaluation metrics across different methods, training corpora, and model architectures. For HoloMotion, the training corpus uses video-derived motions as the primary source of diversity, including motions from MotionMillion~\cite{fan2025go}, and complements them with curated MoCap and in-house motion sources.}
  \label{tab:full_results}
\end{table*}
\FloatBarrier

HoloMotion-1 trained on the full hybrid training corpus of over 2,000 hours consistently achieves the best performance across the evaluation datasets. 
In particular, HoloMotion-1 obtains the lowest mean per-keypoint position error (MPKPE), outperforming the strongest baseline (Sonic) by approximately \textbf{40\%}. 
Improvements are also observed across other metrics including joint position error and root velocity error, indicating more accurate and stable tracking of the reference motions.

Qualitative results further illustrate that HoloMotion-1 is able to reproduce a wider range of motion styles and dynamic behaviors compared to the baseline methods. 
The policy maintains stable tracking under large motion variations and noisy reference signals, which commonly arise in motion reconstructed from in-the-wild videos.

As an integrated release, HoloMotion-1 combines a video-derived hybrid motion corpus with a sparse MoE Transformer policy that has approximately 400M total parameters but activates only about 7M parameters per control step. 
The complete system brings together large-scale heterogeneous motion data, sequence-level training, sparse expert routing, KV-cache inference, and deployment-oriented evaluation, enabling robust zero-shot tracking while keeping inference efficient enough for real-time humanoid control.

\subsection{Training and Inference Efficiency}

The computational efficiency of the proposed sequence-level PPO optimization is evaluated against a traditional single-step sliding-window PPO baseline. 
For a controlled comparison, both methods use the same Dense Transformer policy architecture during profiling.

Training efficiency is measured using 8,192 parallel simulation environments for a single optimization iteration. 
This setup reflects the large-scale parallel training configuration typically used for humanoid control policies. 
Inference latency is evaluated on the robot-side embedded compute module used in our real-world deployment setup.

As illustrated in Figure~\ref{fig:main}-(d), the sequence-level optimization significantly reduces redundant computation associated with overlapping observation prefixes. 
In the training setting, the proposed method achieves approximately a \textbf{22$\times$} improvement in training throughput compared to the sliding-window PPO baseline. 
For inference, the sparse MoE Transformer combined with KV-cache execution reduces policy evaluation latency by approximately \textbf{4$\times$} in the embedded deployment setup.

These results indicate that the proposed training and inference design enables efficient scaling of Transformer-based humanoid control policies while maintaining real-time deployment capability on embedded robotic hardware.

\subsection{Real-World Deployment}

To evaluate sim-to-real transfer, the best-performing HoloMotion-1 policy based on the sparse MoE Transformer architecture is deployed directly on a physical Unitree G1 humanoid robot without any real-world fine-tuning. 
All onboard policy inference runs on the robot-side embedded compute module.

The combination of the sparsely activated MoE architecture and KV-cached Transformer inference enables efficient real-time control. 
The policy is capable of running at approximately 200--300\,Hz on the robot-side embedded platform, while the robot control loop operates at a fixed frequency of 50\,Hz for stable execution.

As illustrated in Figure~\ref{fig:demo}, the trained policy transfers successfully from simulation to the real robot hardware. 
The robot demonstrates robust zero-shot tracking across a wide range of out-of-domain reference motions. 
Examples include highly dynamic dance sequences reconstructed from in-the-wild videos, contact-rich behaviors such as crawling and sitting, and agile martial arts kicking motions.

Additional experiments are conducted using real-time teleoperation devices, including inertial motion capture suits and VR-based tracking systems. 
In these scenarios, the robot follows the human operator’s motion commands with stable tracking and responsive behavior. 
More demonstrations and videos are available on the project page.

\section{Conclusion}

This report introduced \textbf{HoloMotion-1}, a humanoid whole-body motion tracking system trained on a video-derived hybrid motion corpus. 
By combining large-scale video-reconstructed motions with curated MoCap and in-house motion sources, HoloMotion-1 benefits from broad motion diversity while retaining high-quality and deployment-relevant supervision.
The system combines a sparse Mixture-of-Experts (MoE) Transformer architecture with sequence-level policy optimization, enabling efficient training on long motion sequences and high-capacity sequence modeling for whole-body control. 
Together with KV-cached inference, the sparsely activated policy preserves real-time closed-loop execution while scaling model capacity.

Experiments show strong zero-shot generalization across multiple unseen motion datasets spanning diverse activities and capture devices, as well as direct sim-to-real transfer without task-specific fine-tuning. 
These results highlight the value of combining high-capacity sequence models with large-scale heterogeneous motion corpora for scalable humanoid motion control.

\section{Roadmap and Future Work}

\begin{figure*}[htbp]
    \centering
    \includegraphics[width=\textwidth]{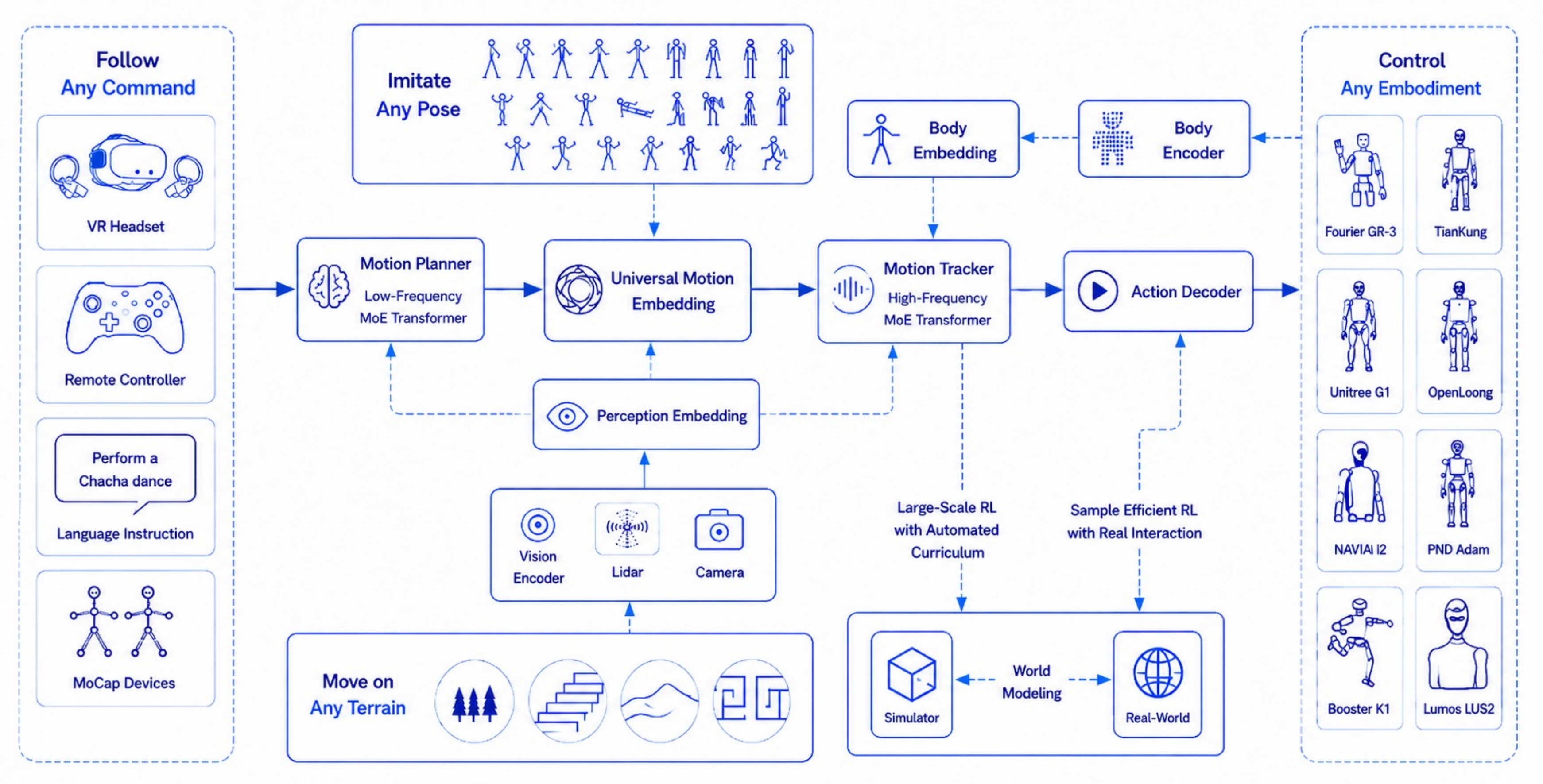}
    \caption{
    Roadmap of HoloMotion toward a foundation model for whole-body humanoid control. 
    HoloMotion-1 establishes the first-stage capability of \textbf{Imitate Any Pose}, while future stages extend the system toward \textbf{Follow Any Command}, \textbf{Move on Any Terrain}, and \textbf{Control Any Embodiment}.
    }
    \label{fig:roadmap}
\end{figure*}

HoloMotion is developed around a 4-Any roadmap for whole-body humanoid control: \textbf{Imitate Any Pose}, \textbf{Follow Any Command}, \textbf{Move on Any Terrain}, and \textbf{Control Any Embodiment}. 
HoloMotion-1 delivers the first stage, \textbf{Imitate Any Pose}, by providing a deployable whole-body tracking policy that can follow diverse reference motions from offline motion clips and online tracking inputs. 
This release also serves as a scalable training and evaluation baseline for users who want to build their own motion tracking models. 
The next stages extend HoloMotion from motion imitation toward command-driven control, terrain-aware interaction, and cross-robot generalization, as summarized in Figure~\ref{fig:roadmap}.

\textbf{Follow Any Command.}
The next stage is command-conditioned humanoid control. 
Instead of requiring a complete full-body reference trajectory at every timestep, future HoloMotion releases will support more compact and user-friendly command interfaces, including velocity commands, remote controllers, partial-body VR or MoCap inputs, and language or task-level instructions. 
The goal is to reuse the real-time stabilization and whole-body coordination learned by HoloMotion-1 while adding a command interface that turns sparse user intent into feasible robot motion.

\textbf{Move on Any Terrain.}
The terrain stage is about making the same whole-body skills usable outside clean flat-floor settings. 
Future HoloMotion releases will extend motion tracking and command following to stairs, slopes, uneven ground, narrow passages, and contact-rich environments. 
This requires terrain-aware perception, such as height maps, depth sensing, and proprioceptive feedback, together with large-scale simulation and automated curriculum learning so that users can deploy motions under more realistic environmental constraints.

\textbf{Control Any Embodiment.}
The embodiment stage targets transfer across different humanoid platforms and robot morphologies. 
Current motion tracking policies are usually tied to one robot's kinematic structure, degrees of freedom, actuator limits, and sensing configuration. 
Future HoloMotion releases will explore morphology-conditioned representations and robot-specific action decoders so that a shared motion foundation model can adapt to humanoids with different proportions, joint layouts, and hardware constraints.

In short, HoloMotion-1 is the \textbf{Imitate Any Pose} release. 
\textbf{Follow Any Command} is the next near-term expansion, while \textbf{Move on Any Terrain} and \textbf{Control Any Embodiment} define longer-term milestones toward general-purpose whole-body humanoid control.

\bibliographystyle{unsrtnat}
\bibliography{paper}

@article{chen2025gmt,
  title={Gmt: General motion tracking for humanoid whole-body control},
  author={Chen, Zixuan and Ji, Mazeyu and Cheng, Xuxin and Peng, Xuanbin and Peng, Xue Bin and Wang, Xiaolong},
  journal={arXiv preprint arXiv:2506.14770},
  year={2025}
}

@inproceedings{mahmood2019amass,
  title={AMASS: Archive of motion capture as surface shapes},
  author={Mahmood, Naureen and Ghorbani, Nima and Troje, Nikolaus F and Pons-Moll, Gerard and Black, Michael J},
  booktitle={Proceedings of the IEEE/CVF international conference on computer vision},
  pages={5442--5451},
  year={2019}
}

@article{harvey2020robust,
  title={Robust motion in-betweening},
  author={Harvey, F{\'e}lix G and Yurick, Mike and Nowrouzezahrai, Derek and Pal, Christopher},
  journal={ACM Transactions on Graphics (TOG)},
  volume={39},
  number={4},
  pages={60--1},
  year={2020},
  publisher={ACM New York, NY, USA}
}

@article{li2023object,
  title={Object motion guided human motion synthesis},
  author={Li, Jiaman and Wu, Jiajun and Liu, C Karen},
  journal={ACM Transactions on Graphics (TOG)},
  volume={42},
  number={6},
  pages={1--11},
  year={2023},
  publisher={ACM New York, NY, USA}
}

@inproceedings{guo2020action2motion,
  title={Action2motion: Conditioned generation of 3d human motions},
  author={Guo, Chuan and Zuo, Xinxin and Wang, Sen and Zou, Shihao and Sun, Qingyao and Deng, Annan and Gong, Minglun and Cheng, Li},
  booktitle={Proceedings of the 28th ACM international conference on multimedia},
  pages={2021--2029},
  year={2020}
}

@inproceedings{fan2025go,
  title={Go to zero: Towards zero-shot motion generation with million-scale data},
  author={Fan, Ke and Lu, Shunlin and Dai, Minyue and Yu, Runyi and Xiao, Lixing and Dou, Zhiyang and Dong, Junting and Ma, Lizhuang and Wang, Jingbo},
  booktitle={Proceedings of the IEEE/CVF International Conference on Computer Vision},
  pages={13336--13348},
  year={2025}
}

@article{luo2025sonic,
  title={Sonic: Supersizing motion tracking for natural humanoid whole-body control},
  author={Luo, Zhengyi and Yuan, Ye and Wang, Tingwu and Li, Chenran and Chen, Sirui and Castaneda, Fernando and Cao, Zi-Ang and Li, Jiefeng and Minor, David and Ben, Qingwei and others},
  journal={arXiv preprint arXiv:2511.07820},
  year={2025}
}

@inproceedings{luo2023perpetual,
  title={Perpetual humanoid control for real-time simulated avatars},
  author={Luo, Zhengyi and Cao, Jinkun and Kitani, Kris and Xu, Weipeng and others},
  booktitle={Proceedings of the IEEE/CVF International Conference on Computer Vision},
  pages={10895--10904},
  year={2023}
}

@article{he2024omnih2o,
  title={Omnih2o: Universal and dexterous human-to-humanoid whole-body teleoperation and learning},
  author={He, Tairan and Luo, Zhengyi and He, Xialin and Xiao, Wenli and Zhang, Chong and Zhang, Weinan and Kitani, Kris and Liu, Changliu and Shi, Guanya},
  journal={arXiv preprint arXiv:2406.08858},
  year={2024}
}

@article{han2025kungfubot2,
  title={Kungfubot2: Learning versatile motion skills for humanoid whole-body control},
  author={Han, Jinrui and Xie, Weiji and Zheng, Jiakun and Shi, Jiyuan and Zhang, Weinan and Xiao, Ting and Bai, Chenjia},
  journal={arXiv preprint arXiv:2509.16638},
  year={2025}
}

@article{zhang2025track,
  title={Track any motions under any disturbances},
  author={Zhang, Zhikai and Guo, Jun and Chen, Chao and Wang, Jilong and Lin, Chenghuai and Lian, Yunrui and Xue, Han and Wang, Zhenrong and Liu, Maoqi and Lyu, Jiangran and others},
  journal={arXiv preprint arXiv:2509.13833},
  year={2025}
}

@article{vaswani2017attention,
  title={Attention is all you need},
  author={Vaswani, Ashish and Shazeer, Noam and Parmar, Niki and Uszkoreit, Jakob and Jones, Llion and Gomez, Aidan N and Kaiser, {\L}ukasz and Polosukhin, Illia},
  journal={Advances in neural information processing systems},
  volume={30},
  year={2017}
}

@article{radford2019language,
  title={Language models are unsupervised multitask learners},
  author={Radford, Alec and Wu, Jeffrey and Child, Rewon and Luan, David and Amodei, Dario and Sutskever, Ilya and others},
  journal={OpenAI blog},
  volume={1},
  number={8},
  pages={9},
  year={2019}
}

@article{brown2020language,
  title={Language models are few-shot learners},
  author={Brown, Tom and Mann, Benjamin and Ryder, Nick and Subbiah, Melanie and Kaplan, Jared D and Dhariwal, Prafulla and Neelakantan, Arvind and Shyam, Pranav and Sastry, Girish and Askell, Amanda and others},
  journal={Advances in neural information processing systems},
  volume={33},
  pages={1877--1901},
  year={2020}
}

@inproceedings{zitkovich2023rt,
  title={Rt-2: Vision-language-action models transfer web knowledge to robotic control},
  author={Zitkovich, Brianna and Yu, Tianhe and Xu, Sichun and Xu, Peng and Xiao, Ted and Xia, Fei and Wu, Jialin and Wohlhart, Paul and Welker, Stefan and Wahid, Ayzaan and others},
  booktitle={Conference on Robot Learning},
  pages={2165--2183},
  year={2023},
  organization={PMLR}
}

@article{fu2024humanplus,
  title={Humanplus: Humanoid shadowing and imitation from humans},
  author={Fu, Zipeng and Zhao, Qingqing and Wu, Qi and Wetzstein, Gordon and Finn, Chelsea},
  journal={arXiv preprint arXiv:2406.10454},
  year={2024}
}

@article{tessler2024maskedmimic,
  title={Maskedmimic: Unified physics-based character control through masked motion inpainting},
  author={Tessler, Chen and Guo, Yunrong and Nabati, Ofir and Chechik, Gal and Peng, Xue Bin},
  journal={ACM Transactions On Graphics (TOG)},
  volume={43},
  number={6},
  pages={1--21},
  year={2024},
  publisher={ACM New York, NY, USA}
}

@article{radosavovic2024humanoid,
  title={Humanoid locomotion as next token Prediction. arXiv 2024},
  author={Radosavovic, I and Zhang, B and Shi, B and Rajasegaran, J and Kamat, S and Darrell, T and Sreenath, K and Malik, J},
  journal={arXiv preprint arXiv:2402.19469},
  year={2024}
}

@article{bi2025h,
  title={H-rdt: Human manipulation enhanced bimanual robotic manipulation},
  author={Bi, Hongzhe and Wu, Lingxuan and Lin, Tianwei and Tan, Hengkai and Su, Zhizhong and Su, Hang and Zhu, Jun},
  journal={arXiv preprint arXiv:2507.23523},
  year={2025}
}

@article{wang2025experts,
  title={From experts to a generalist: Toward general whole-body control for humanoid robots},
  author={Wang, Yuxuan and Yang, Ming and Ding, Ziluo and Zhang, Yu and Zeng, Weishuai and Xu, Xinrun and Jiang, Haobin and Lu, Zongqing},
  journal={arXiv preprint arXiv:2506.12779},
  year={2025}
}

@article{zhang2019root,
  title={Root mean square layer normalization},
  author={Zhang, Biao and Sennrich, Rico},
  journal={Advances in neural information processing systems},
  volume={32},
  year={2019}
}

@article{su2024roformer,
  title={Roformer: Enhanced transformer with rotary position embedding},
  author={Su, Jianlin and Ahmed, Murtadha and Lu, Yu and Pan, Shengfeng and Bo, Wen and Liu, Yunfeng},
  journal={Neurocomputing},
  volume={568},
  pages={127063},
  year={2024},
  publisher={Elsevier}
}

@inproceedings{ainslie2023gqa,
  title={Gqa: Training generalized multi-query transformer models from multi-head checkpoints},
  author={Ainslie, Joshua and Lee-Thorp, James and De Jong, Michiel and Zemlyanskiy, Yury and Lebr{\'o}n, Federico and Sanghai, Sumit},
  booktitle={Proceedings of the 2023 Conference on Empirical Methods in Natural Language Processing},
  pages={4895--4901},
  year={2023}
}

@inproceedings{henry2020query,
  title={Query-key normalization for transformers},
  author={Henry, Alex and Dachapally, Prudhvi Raj and Pawar, Shubham Shantaram and Chen, Yuxuan},
  booktitle={Findings of the Association for Computational Linguistics: EMNLP 2020},
  pages={4246--4253},
  year={2020}
}

@article{qiu2025gated,
  title={Gated attention for large language models: Non-linearity, sparsity, and attention-sink-free},
  author={Qiu, Zihan and Wang, Zekun and Zheng, Bo and Huang, Zeyu and Wen, Kaiyue and Yang, Songlin and Men, Rui and Yu, Le and Huang, Fei and Huang, Suozhi and others},
  journal={arXiv preprint arXiv:2505.06708},
  year={2025}
}

@article{schulman2017proximal,
  title={Proximal policy optimization algorithms},
  author={Schulman, John and Wolski, Filip and Dhariwal, Prafulla and Radford, Alec and Klimov, Oleg},
  journal={arXiv preprint arXiv:1707.06347},
  year={2017}
}

@article{ze2025twist2,
  title={Twist2: Scalable, portable, and holistic humanoid data collection system},
  author={Ze, Yanjie and Zhao, Siheng and Wang, Weizhuo and Kanazawa, Angjoo and Duan, Rocky and Abbeel, Pieter and Shi, Guanya and Wu, Jiajun and Liu, C Karen},
  journal={arXiv preprint arXiv:2511.02832},
  year={2025}
}

\end{document}